\documentclass[10pt,twocolumn,letterpaper]{article}

\usepackage[pagenumbers]{cvpr} 

\usepackage{amsmath}  
\usepackage{graphicx}
\usepackage{bm}
\usepackage{multirow}
\usepackage{makecell}
\usepackage{cuted}
\usepackage{capt-of}
\usepackage[hypcap=false]{caption}
\definecolor{cvprblue}{rgb}{0.21,0.49,0.74}
\usepackage[pagebackref,breaklinks,colorlinks,allcolors=cvprblue]{hyperref}
\newcommand{\myparagraph}[1]{\vspace{0.1em}\noindent\textbf{#1}}

\title{InterAgent: Physics-based Multi-agent Command Execution via Diffusion on Interaction Graphs}

\author{
Bin Li$^*$ \textsuperscript{1} \quad 
Ruichi Zhang$^*$ \textsuperscript{2} \quad 
Han Liang \textsuperscript{\textdagger} \textsuperscript{3} \quad 
Jingyan Zhang \textsuperscript{1} \quad 
Juze Zhang \textsuperscript{4} \quad \\
Xin Chen \textsuperscript{3} \quad 
Lan Xu \textsuperscript{1} \quad 
Jingyi Yu \textsuperscript{1} \quad 
Jingya Wang \textsuperscript{\textdagger} \textsuperscript{1, 5} \\
\textsuperscript{1} \textit{ShanghaiTech University} \quad 
\textsuperscript{2} \textit{University of Pennsylvania} \quad 
\textsuperscript{3} \textit{ByteDance} \quad \\
\textsuperscript{4} \textit{Stanford University} \quad
\textsuperscript{5} \textit{InstAdapt}\\
}

\begin{document}
\maketitle
\renewcommand{\thefootnote}{\fnsymbol{footnote}}
\footnotetext[1]{Equal contribution.}
\footnotetext[2]{Corresponding authors.}
\begin{abstract}

Humanoid agents are expected to emulate the complex coordination inherent in human social behaviors. However, existing methods are largely confined to single-agent scenarios, overlooking the physically plausible interplay essential for multi-agent interactions. To bridge this gap, we propose InterAgent, the first end-to-end framework for text-driven physics-based multi-agent humanoid control. At its core, we introduce an autoregressive diffusion transformer equipped with multi-stream blocks, which decouples proprioception, exteroception, and action to mitigate cross-modal interference while enabling synergistic coordination. We further propose a novel interaction graph exteroception representation that explicitly captures fine-grained joint-to-joint spatial dependencies to facilitate network learning. Additionally, within it we devise a sparse edge-based attention mechanism that dynamically prunes redundant connections and emphasizes critical inter-agent spatial relations, thereby enhancing the robustness of interaction modeling. Extensive experiments demonstrate that InterAgent consistently outperforms multiple strong baselines, achieving state-of-the-art performance. It enables producing coherent, physically plausible, and semantically faithful multi-agent behaviors from only text prompts. Our code and data will be released to facilitate future research.
Project page: \tt \small \textbf{\href{https://binlee26.github.io/InterAgent-Page}{https://binlee26.github.io/InterAgent-Page}}.

\end{abstract}    
\section{Introduction}\label{sec:intro}
    
    

\begin{figure}[tpb]
    \centering

    \includegraphics[width=\columnwidth]{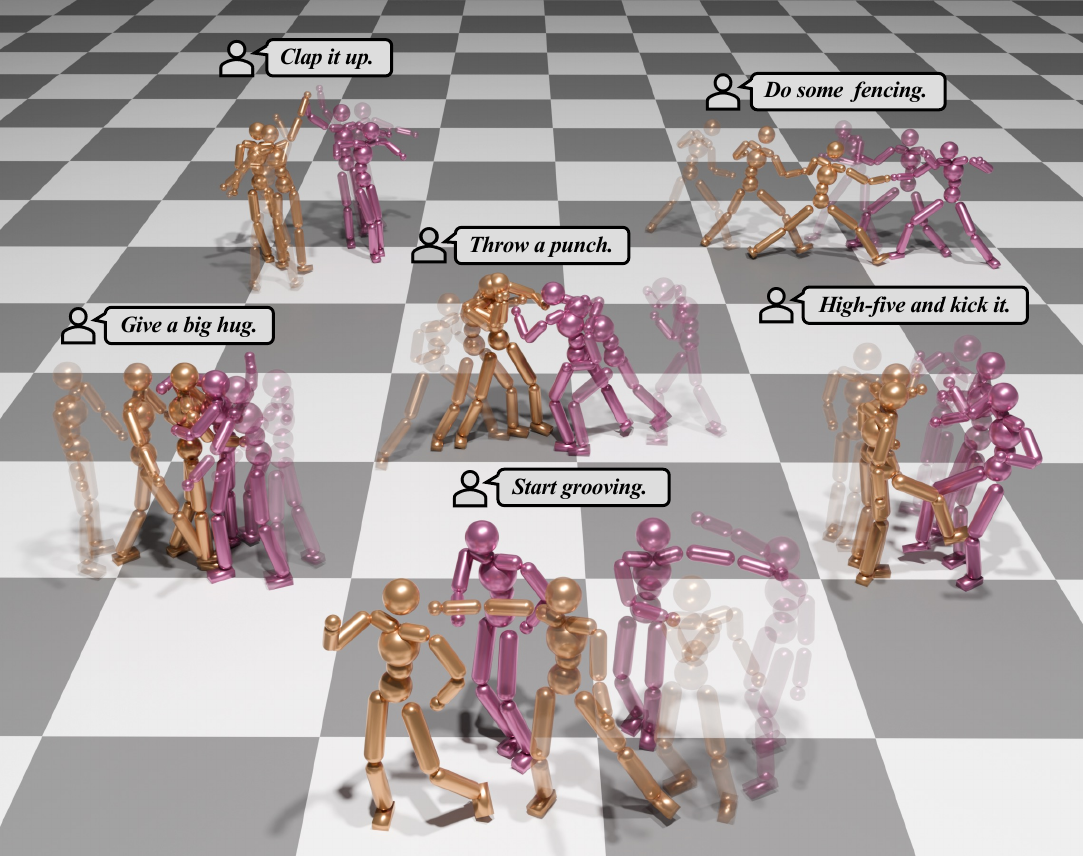}
    
    \caption{
    \textbf{InterAgent} produces physically plausible multi-agent interactions across diverse scenarios from only text prompts.
    }
    \label{fig:teaser}
\end{figure}

Humanoid agents should embody the intrinsic ways we human beings interact and communicate with each other, so as to depict the diverse cultures and societies that make up our physical world.
Empowering such agents to perform coordinated behaviors in shared environments stands as a core computer vision challenge, with various applications spanning gaming, VR/AR, and autonomous robots.

Recent kinematics-based approaches for human motion generation, empowered by generative models such as diffusion models~\citep{liang2024intergen,cai2024digital,xu2024inter,ruiz2024in2in,fan2024freemotion,wang2024intercontrol} and autoregressive models~\citep{javed2024intermask}, have demonstrated promising capabilities in synthesizing interactive motions.
Yet, they neglect physical plausibility, often introducing artifacts such as body penetration, floating, or unnatural sliding—significantly compromising both the realism and practical applications.
In parallel, the rapid advancement of reinforcement learning and physics simulation has spurred growing interest in physics-based methods~\citep{rocca2025policy1,rocca2025policy2,tevet2024closd,truong2024pdp,wu2025uniphys,huang2024diffuseloco,zhang2025add,cui2025grove}.
Typically, the generation-and-track frameworks like PhysDiff~\cite{yuan2023physdiff} and CLoSD~\citep{tevet2024closd} leverage imitation-learned motion tracking policy to project generated kinematic motions into physically plausible trajectories that minimize tracking error.
Yet, they struggle with the inherent discrepancy between kinematic priors and physics-based tracking, often producing unexpected falling movements.
Alternatively, recent diffusion policy techniques~\cite{chi2023diffusion,ren2024diffusion,wu2025afforddp,zhu2025scaling,xue2025reactive} enables training end-to-end generative policies, such as PDP~\citep{truong2024pdp}, UniPhys~\citep{wu2025uniphys}, and Diffuse-CLoC~\citep{huang2024diffuseloco}.
They form a track-then-distill paradigm, in which a tracking policy is first trained to collect physically plausible state-action trajectories from human motion dataset, which are subsequently used to train generative model as behavior cloner in an end-to-end fashion.
Despite their effectiveness, these approaches are predominantly confined to single-agent setting, leaving the rich and complex dynamics of multi-agent interactions—such as collaborative tasks or social-aware behaviors largely underexplored.

To bridge this gap, we introduce InterAgent, a novel physics-based end-to-end framework for text-driven multi-agent humanoid control, which enables producing vivid and physically plausible multi-agent interactions from only text prompts.
Specifically, in our multi-agent systerm, each agent governs its own dynamics through proprioception while perceiving others via exteroception. However, modeling exteroception naively as relative dynamical states overlooks the fine-grained joint-to-joint dependencies crucial for coordinated interactions. To overcome this limitation, we propose to incorporate \textbf{interaction graph exteroception} representation into diffusion framework, wherein joints are modeled as nodes interconnected via directed edges. This formulation enables explicit and granular modeling of inter-agent relations.
Motivated by the inherent modality heterogeneity among proprioception, exteroception, and action, our Interaction Diffusion Transformer (Inter-DiT) employs a multi-stream architecture that treats each modality as a distinct yet complementary stream. Each \textbf{multi-stream DiT block} integrates an inter-stream fusion attention to facilitate cross-modal information exchange, alongside a context-aware conditioning attention that incorporates temporal and inter-agent contextual cues. This design preserves the unique characteristics of each modality while enabling synergistic coordination, thereby achieving a more holistic representation of interaction dynamics.
Furthermore, inspired by the sparse and selective nature of real-world human interactions, we introduce a tailored \textbf{edge-based sparse attention} mechanism on interaction graph within the exteroception stream. This mechanism dynamically suppresses edges that contribute minimally to the ongoing interaction, thereby focusing the model’s attention on the most salient joint-level dependencies and enhancing the plausibility of synthesized interactions.
Extensive experiments demonstrate that our approach consistently outperforms several baselines, generating interactive behaviors that are physically plausible, semantically aligned, and rich in fine-grained coordination.

To summarize, our main contributions include: 
\begin{itemize}
    \item We propose InterAgent, the first end-to-end framework for physics-based text-driven multi-agent humanoid control, achieving state-of-the-art performance.

    \item We propose Inter-DiT, which incorporates a novel multi-stream DiT block design that decouples proprioception, exteroception and action modeling, effectively improving the interaction results.

    \item We propose interaction graph exteroception representation and a tailored sparse edge-based attention mechanism, which further improves the robustness of inter-agent spatial dependencies.
    
\end{itemize}

\section{Related Work}\label{sec:related work}
 
\myparagraph{Physics-based humanoid control.}
A long-standing goal in computer animation is to design controllers that produce agile and life-like behaviors. Early optimization-based methods~\citep{coros2010generalized,hodgins1995animating,yin2007simbicon,levine2013guided} explicitly crafted skill-specific objectives but relied on labor-intensive heuristics and lacked generalization. Recent advances increasingly adopt reinforcement learning to learn controllers from data. Many works focus on motion tracking, where policies imitate reference clips~\citep{peng2018deepmimic,liu2018learning,luo2021dynamics,Luo2023PerpetualHC,luo2023universal,simos2025reinforcement,huang2025modskill,wang2025hil,fussell2021supertrack,wu2024cbil,he2025asap,wang2024pacer+,yin2025unitracker}. Among them, PHC~\citep{Luo2023PerpetualHC} and PULSE~\citep{luo2023universal} scaling tracking to large datasets. Beyond tracking, other studies~\citep{peng2021amp,peng2022ase,yao2022controlvae,wang2020unicon,zhu2023neural} learn general motion priors from unstructured motion data, enabling agents to compose and adapt skills.
Text-conditioned controllers~\citep{juravsky2022padl,ren2023insactor,truong2024pdp,juravsky2024superpadl,yao2024moconvq,cui2024anyskill,cui2025grove} further map language to actions, as in PADL~\citep{juravsky2022padl} and PDP~\citep{truong2024pdp}, but remain limited to single-agent scenarios.
To move beyond, recent works explore physics-based synthesis of human–object interactions~\citep{akkerman2025interdyn,cao2024multi,christen2024diffh2o,xu2023interdiff,peng2025hoi,tessler2025maskedmanipulator,wang2024strategy,pan2025tokenhsi,yu2025skillmimic,wang2025skillmimic,kim2025physicsfc}, such as InterMimic~\citep{xu2025intermimic} and CooHOI~\citep{gao2024coohoi}, enable single-agent skill coordination with dynamic objects but still struggle with complex multi-agent interactions.

\myparagraph{Human interaction synthesis.}
Realistic interaction behaviors, where agents coordinate and respond to each other, have attracted increasing attention. Most approaches are kinematic and data-driven~\citep{xu2024inter,chopin2023interaction,xu2024regennet,ruiz2024in2in,fan2024freemotion,shafir2023human,tanaka2023role,wu2025text2interact,liu2024physreaction,yao2025physiinter,yi2024generating,cen2025ready}, such as InterGen~\citep{liang2024intergen} with a diffusion-based text-to-interaction model, and Think-Then-React~\citep{tan2025think} leveraging large language models for joint intention inference and motion prediction. While effective for semantic alignment, these methods ignore physical feasibility and dynamic interactions. Physics-based approaches~\citep{won2021control,zhang2023simulation} produce physically plausible behaviors and capture interaction forces, but remain limited to task-specific settings or rely on a narrow set of reference motions, restricting their ability to model diverse interactive behaviors.

\myparagraph{Diffusion models in physics-based humanoid control.}
Diffusion models~\citep{sohl2015deep,ho2020denoising,yang2023diffusion,nichol2021improved,dhariwal2021diffusion,song2020denoising} have recently emerged as powerful generative frameworks for control, capturing multimodal action distributions and flexible conditioning~\citep{tevet2022human,tseng2023edge,huang2024diffuseloco,janner2022planning,yuan2023physdiff,chi2025diffusion,chen2024taming,shi2024interactive,rempe2023trace}. CLoSD~\citep{tevet2024closd} separates a diffusion motion planner from a tracking controller, while PDP~\citep{truong2024pdp} learns an end-to-end diffusion policy for direct control. Diffuse-CLoC~\citep{huang2025diffuse} and UniPhys~\citep{wu2025uniphys} further unify state–action modeling and long-horizon planning. Despite these advances, the existing diffusion-based frameworks are restricted to single-agent control, and the complex multi-agent interaction setting remains unsolved.
\section{Preliminaries} \label{sec:prelim}
\myparagraph{Physics simulation setup.} We simulate the dynamics of physics-based humanoids~\citep{peng2018deepmimic,peng2021amp} in Isaac Gym~\citep{liang2018gpu}. Each humanoid agent is equipped with 15 joints and 28 actuators, with each actuator controlled via a proportional-derivative (PD) controller, and the action $\mathbf{a}_t$ specifies the target inputs for its actuators. Given the current states $\mathbf{s}_t$, the physics simulator advances the humanoids to the next states $\mathbf{s}_{t+1}$, sampled according to  the dynamic transition $\mathbf{s}_{t+1} \sim p(\mathbf{s}_{t+1} | \mathbf{s}_{t}, \mathbf{a}_t) $.

\myparagraph{Interaction tracking policy.}
Previous works~\cite{peng2018deepmimic,Luo2023PerpetualHC,luo2023universal} employ reinforcement learning to replicate MoCap datasets, which effectively imitate the reference motions with a learned tracking policy, $\bm{a}_t = \bm{\pi}(\bm{s}_{t}, \bm{s}_{t+1}^{ref})$, where $\bm{s}_{t+1}^{ref}$ is the next-frame reference state from reference motions.
We adopt a curriculum learning strategy analogous to PHC~\citep{Luo2023PerpetualHC}, which progresses from simpler to more complex motions. 
To faithfully capture the dynamics of inter-agent interactions, we supplement the tracking reward with an interaction graph reward~\citep{zhang2023simulation}—a design that explicitly enforces the natural spatial relationships between interacting agents.

\myparagraph{Data collection.}
Next, we leverage the trained expert policies to collect state-action trajectories, constituting the tracking dataset. To enhance state space coverage, we augment trajectory diversity by noise disturbance, as introduced in PDP~\citep{truong2024pdp}. Specifically, for a given motion sequence $\bm m = \{{\bm s}_t^{ref}\}_{t=1}^n$, we deploy its corresponding tracking policy $\bm{\pi}$ in the simulator to roll out diverse trajectories like $\{{\bm s}_1, {\bm a}_1, ..., {\bm s}_n, {\bm a}_n\}$, where ${\bm a}_t = \bm{\pi}({\bm s}_t,{\bm s}_t^{ref}) + {\epsilon}$ with ${\epsilon} \sim \mathcal{N}(0, {\sigma}^2)$. Finally, the noisy-state clean-action trajectory $\{{\bm s}_1, \bm{\pi}({\bm s}_1,{\bm s}_1^{ref}), ..., {\bm s}_n, \bm{\pi}({\bm s}_n,{\bm s}_n^{ref})\}$, along with the text description associated with the motion sequence $\bm m$, are collected into the dataset $\mathcal{D}$. In practice, to ensure the reliability of the collected data, we perform multiple rollouts for each motion and select 8 successful trajectories as training samples. Empirically, a noise level of $\sigma=0.01$ strikes a good balance between bandwidth and temporal stability.

\section{Method} \label{sec:method}
\begin{figure}
    \centering

    \includegraphics[width=\columnwidth]{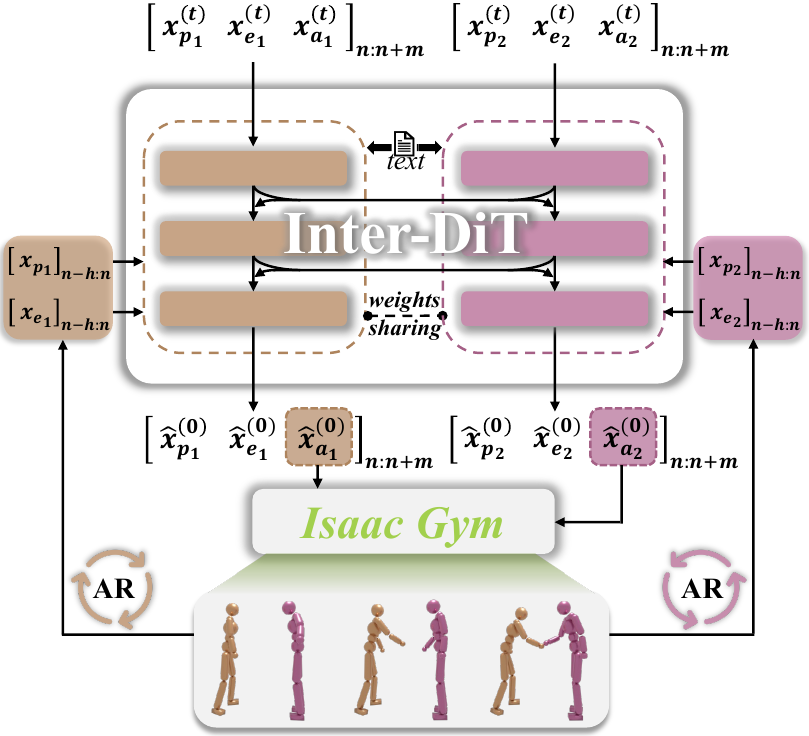}
    
    \caption{
    \textbf{InterAgent overview.} 
 A physics-based framework for text-driven multi-agent interactive behavior generation, built upon Inter-DiT — two cooperative, weight-sharing networks under an autoregressive diffusion paradigm.
    }
    \label{fig:overview_pipeline}
\end{figure}

We focus on physics-based text-driven control for two interacting humanoid agents. In single-agent settings, \textbf{actions} depend mainly on an agent’s own state (\textbf{proprioception}). In contrast, multi-agent scenarios introduce additional challenges: each agent’s actions are influenced not only by its intrinsic dynamics but also by the other’s states and behaviors (\textbf{exteroception}).
To this end, we propose a novel framework InterAgent (Fig.~\ref{fig:overview_pipeline}). It incorporates an Interaction Diffusion Transformer (Inter-DiT), composed of two cooperative, weight-sharing networks under an autoregressive diffusion paradigm (\cref{sec:idt}), to effectively model interactive dynamics.
Given the inherent heterogeneity among proprioception, exteroception, and action, we treat them as distinct modalities. To handle these modalities in a coordinated manner, Inter-DiT adopts a multi-stream architecture (\cref{sec:block}) that enables decoupled yet cooperative modeling and enhances overall performance. Moreover, we propose a novel and effective exteroception representation, interaction graph (IG), and devise a tailored edge-based sparse attention mechanism on the exteroception stream (\cref{sec:sig}) to selectively suppress interaction-irrelevant connections and effectively highlight salient inter-agent relations, based on its sparsity nature.
\begin{figure}[htbp]
    \centering
    
    \includegraphics[width=\columnwidth]{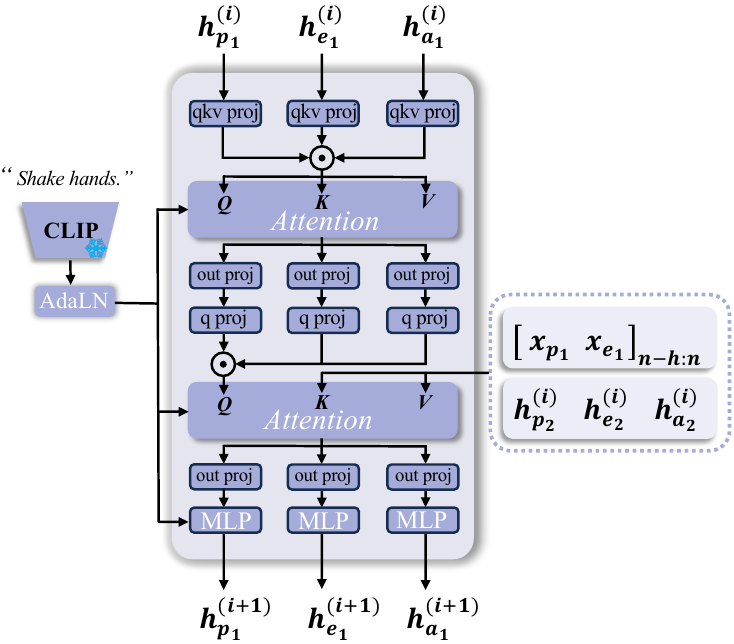}
\caption{\textbf{Multi-stream DiT block design.} Each modality—proprioception, exteroception, and action—is processed in an independent stream. Inter-stream fusion attention exchanges information across modalities, while context-aware conditioning attention integrates temporal and inter-agent context, enabling decoupled yet coordinated modeling.
    }
    \label{fig:interdit}
\end{figure}

\subsection{Interaction Diffusion Transformer}\label{sec:idt} 

We propose the Interaction Diffusion Transformer (Inter-DiT), a novel autoregressive diffusion framework designed to model the interactive behaviors between agents, as illustrated in \cref{fig:overview_pipeline}. Inspired by Diffuse-CLOC~\citep{huang2024diffuseloco} and Uniphys~\citep{wu2025uniphys}, we model the joint distribution of states and actions under textual conditions $\mathbf{c}$, enabling coherent prediction of future state-action pairs with implicitly learned dynamic transition world model. 

\myparagraph{Humanoid behavior representation.} 
We represent the humanoid behavior sequence as $\bm{X} = \bm{x}_{1:T} =  [ \bm{x}_s, \bm{x}_a ]_{1:T} $, where $\bm{x}_s$ and $\bm{x}_a$ denotes the state and action, respectively. To be specific, each state  $\bm{x}_s$ is composed of the following elements:
\begin{equation}\label{state}
    \bm{x}_s = (R_h, \bm{p}, \bm{r}, \bm{v}, \bm{w}),
\end{equation}

where $R_h \in \mathbb{R}$ denotes the heights of the root from the ground. $\bm{p}\in \mathbb{R}^{(J-1) \times 3}$ denotes the 3D position of each joint (excluding the root) within the humanoid's root frame. $\bm{r}\in \mathbb{R}^{J \times 6}$ encodes the local rotation of each joint using 
6D representation~\citep{zhou2019continuity}. In addition, $\bm{v}\in \mathbb{R}^{J \times 3}$ and $\bm{w}\in \mathbb{R}^{J \times 3}$ is the local linear and angular velocities of each joint, respectively. 
As for each action $\bm{x}_a$, it specifies the target joint rotations used by PD controllers to drive the humanoid's movements. The resulting control signals form a 28-dof action space for each humanoid.

\myparagraph{Interaction graph exteroception.} 
In our multi-agent scenarios, the behavior of each agent is inherently influenced by its spatiotemporal interactions with others. To capture such interdependencies, we extend the state representation $\bm{x}_s$ in Eq.~\eqref{state} by incorporating an additional exteroception component, yielding $\bm{x}_s = [\bm{x}_p, \bm{x}_e]$, where $\bm{x}_p$ represents the proprioception (as defined in Eq.~\eqref{state}) and $\bm{x}_e$ encodes exteroceptive information reflecting inter-agent interactions.
An intuitive design for exteroception modeling is to directly represent the other agent's proprioception state in the agent's own root frame, which we refer to as relative state (RS) exteroception.

\begin{figure}[tpb]
    \centering

    \includegraphics[width=\columnwidth]{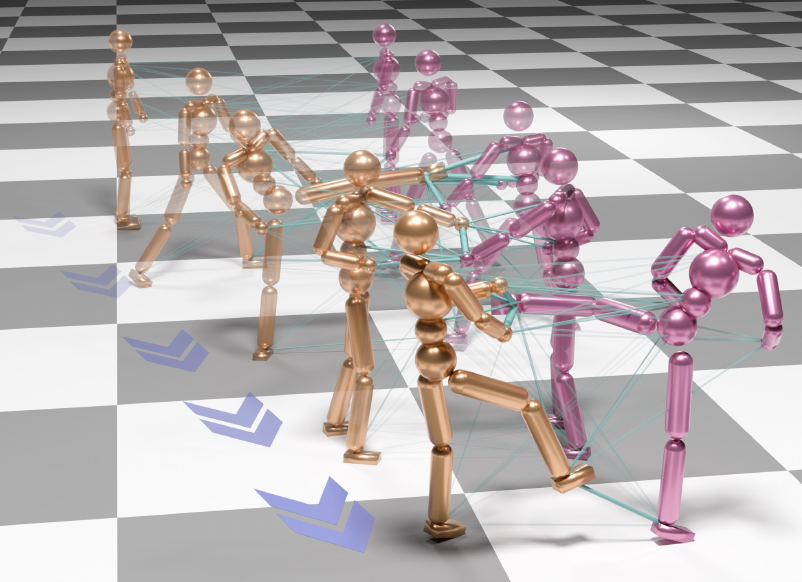}
    
    \caption{
    \textbf{Sparse Interaction Graph.} 
    Each joint of one character connects to all joints of the other via directed edges (dark green), where each edge vector encodes the spatial interaction between the corresponding joints. The thickness of each edge encodes the magnitude of its contribution to the interactive dynamics. Light purple arrows on the ground indicate temporal progression. 
    }
    \label{fig:intergaph}
\end{figure}

However, we find that such a naive exteroception representation is suboptimal for learning complex interactive patterns.
Drawing inspiration from~\citet{zhang2023simulation}, we propose the interaction graph (IG) as the exteroception, which provides a more explicit and structured representation of inter-agent dynamics. Specifically, for each joint position $\bm{p}_j\in \mathbb{R}^3$ of a humanoid, we construct directed edges towards every joint position $\bm{p}_i\in \mathbb{R}^3$ of the other humanoid, where each edge $\bm{e}_{ij} = \bm{p}_i - \bm{p}_j \in \mathbb{R}^3 $ encodes the spatial interaction between the corresponding joints, as shown in \cref{fig:intergaph}. We refer to this as fully connected IG (FIG) exteroception, written as
\begin{equation}{\label{FIG extra}}
\bm{x}_{e} = (\bm{e}_{1,1},\bm{e}_{1,2},...,\bm{e}_{J,J}) \in \mathbb{R}^{(J*J) \times 3},
\end{equation}
where $J$ denotes the number of joints of one humanoid. 

\myparagraph{Training and inference.}
For simplicity, we omit the explicit agent indices and use a single notation to represent both agents. 
As illustrated in \cref{fig:overview_pipeline}, Inter-DiT models the symmetry of two agents using two cooperative and weight-sharing networks, inspired by InterGen~\citep{liang2024intergen}.
It takes as input a noisy interaction behavior sequence $\bm{X}^{(t)} = [\bm{x}^{(t)}_{p},\bm{x}^{(t)}_{e},\bm{x}^{(t)}_{a}]_{n:n+m}$ of length $m$ and output a denoised sequence. The past $h$ frames of proprioception and exteroception, $\bm{S}=[\bm{x}_{p}, \bm{x}_{e}]_{n-h:n}$, serve as historical states over the previous $h$ frames, where $n$ denotes the current frame.  
Inter-DiT is trained to predict the future behaviors $\bm{\hat{X}}^{(0)}$ given all agents' historical behavior states $\bm{S}$ and text condition $\bm{c}$ in an autoregressive manner. The training objective is defined as:
\begin{equation}
    \mathcal{L} = \mathbb{E}_{t,\bm{X}}[||\bm{X} - \Phi(\bm{X}^{(t)},t,\mathbf{c},\bm{S} )|| ],
\end{equation}
where $t$ is the diffusion timestep. 

During inference, states from the simulator are stored in a first-in-first-out (FIFO) history buffer. Given a command $\bm{c}$, Inter-DiT starts from a noisy sequence, uses the latest $h$ frames as context, and predicts actions $\bm{\hat{x}}_a$ to drive the humanoids, repeating autoregressively for the desired steps.

\subsection{Multi-stream DiT Block}\label{sec:block} 
As mentioned, Inter-DiT enables coherent prediction of future state-action pairs. Unlike prior works~\citep{huang2024diffuseloco, wu2025uniphys} that merge states and actions into a single representation, we treat them as distinct modalities of behavior to reduce inter-modal interference. In multi-agent scenarios, the proprioception and exteroception should also be considered as separate modalities due to their heterogeneous nature. Motivated by prior works~\citep{esser2024scaling,won2025dual,liu2025separate,liu2023dual}, Inter-DiT employs multi-stream blocks to model these distinct modalities as disentangled streams. As depicted in \cref{fig:interdit}, each multi-stream Inter-DiT block is structured into inter-stream fusion attention and contextual conditioning attention stages.

\myparagraph{Inter-stream fusion attention.}
Firstly, the $i$-th hidden features of an agent $[\bm{h}^{(i)}_{p_1}$, $\bm{h}^{(i)}_{e_1}$, $\bm{h}^{(i)}_{a_1}]$ are independently projected into a shared feature space. Then, the features from each stream are concatenated along the sequence dimension, and the resulting aggregated features are encoded via self-attention. After attention, the outputs are then split and routed through dedicated projection layers to mitigate inter-modal interference.

\myparagraph{Context-aware conditioning attention.}
To infuse temporal and inter-agent contextual cues, the outputs from all streams of the previous stage are concatenated along the sequence dimension and used as queries. The keys and values are provided sequentially from history observations $[\bm{x}_{p_1}, \bm{x}_{e_1}]_{n-h:n}$, and the hidden features $[\bm{h}^{(i)}_{p_2}$, $\bm{h}^{(i)}_{e_2}$, $\bm{h}^{(i)}_{a_2}]$ of the other agent. After sequentially attending to each contextual cue with a dedicated attention layer~\citep{vaswani2017attention}, the resulting outputs are split and passed through their respective projection layers.
Finally, the output of each stream is processed by a feed-forward MLP to yield the $(i+1)$-th hidden features $[\bm{h}^{(i+1)}_{p_1}$, $\bm{h}^{(i+1)}_{e_1}$, $\bm{h}^{(i+1)}_{a_1}]$. In addition, an adaptive layer normalization~\citep{peebles2023scalable} is introduced before each attention and feed-forward layer to incorporate conditioning on the textual signal $\bm{c}$ and the timestep $t$. 

\subsection{Sparse Interaction Graph}\label{sec:sig}

Furthermore, we argue that the FIG exteroception $\bm{x}_e$ built from all $J$ joints of each humanoid contradicts the inherently sparse nature of inter-agent relations. For instance, during a handshake, the interaction is primarily governed by the agents's arms and hands, whereas the lower-body joints contribute minimally to the exchange of interactive behavior dynamics. This indicates that the model should selectively focus on the hand-to-hand relations, where salient motion cues and inter-agent dependencies emerge.

\myparagraph{Sparse attention on interaction graph.} To this end, we incorporate a sparse attention mechanism to suppress redundant relations in the FIG exteroception and selectively enhance the most salient inter-agent dependencies. Specifically, the FIG sequence contains $l_e*J*J$ edges, where $l_e$ is the FIG sequence length. These edges are evenly partitioned into subsets, each handled by a distinct attention head.
Each head computes the attention logit between its corresponding queries and the assigned edges.  Mathematically, a feature sequence $\bm{f}\in \mathbb{R}^{l_f \times d_f}$ is input as a query source, where $l_f$ denotes the length of the feature sequence, $d_f$ represents the dimension of the features. The query projection layer projects $\bm{f}$ to a query $\bm{Q} = \bm{f} \cdot \bm{W}^Q$, where $\bm{W}^Q \in \mathbb{R}^{d_f \times d_f}$. And the reshaped FIG edges $\bm{x}_{e} \in \mathbb{R}^{(l_e*J*J) \times 3}$ are projected to a key $\bm{K} = \bm{x}_{e}\cdot\bm{W}^K$, and a value $\bm{V} = \bm{x}_{e} \cdot \bm{W}^V$, where $\bm{W}^K, \bm{W}^V \in \mathbb{R}^{3 \times d_f}$. Then we obtain the attention map $\bm{A} \in \mathbb{R}^{l_f \times (l_e*J*J)}$ with the Gumble-Softmax~\citep{dang2022diverse,yang2019modeling}:
\begin{equation}\label{logits}
   \bm{A}  = \text{Gumble-Softmax}(\frac{\bm{Q}\bm{K}^T}{\sqrt{d_f}}).
\end{equation}
Next, we derive a binary mask $\bm{M}$ to retain the most significant edges with the high scores, while filtering out the rest to enforce structural sparsity.
\begin{equation}\label{logits mask}
    \bm{M}_{ij} = 
    \begin{cases}
       1 ,& j \in \arg \text{TopK}_k(\bm{A}_i)\\
        0 ,& \text{otherwise}\\
    \end{cases},
\end{equation}
where $\arg \text{TopK}_k(\cdot)$ denotes the indices corresponding to the top-$k$ columns of $\bm{A}_i$ ranked by its values, $\bm{A}_i$ is the $i$-th row of $\bm{A}$. Finally, we can get the output $\bm{f}'$ with the following attention mechanism:
\begin{equation}\label{sparse attn}
    \bm{f}' = (\bm{M} \circ \bm{A} ) \bm{V},
\end{equation}
where $\circ$ denotes the Hadamard product. To highlight the inherently sparse structure, the full connected IG in Eq.~\eqref{FIG extra}, after processing through Eqs. \eqref{logits},~\eqref{logits mask}, and ~\eqref{sparse attn}, is denoted as the Sparse IG (SIG) for clarity.

\section{Experiments} \label{sec:experiments}
\subsection{Experiment Setup}

\begin{table*}
\centering
\resizebox{1.0\textwidth}{!}{
\begin{tabular}{cccccccc}
\toprule
&\multicolumn{3}{c}{R-precision $\uparrow$} &  \multirow{2}{*}{FID $\downarrow$} & \multirow{2}{*}{MM dist $\downarrow$} & \multirow{2}{*}{Diversity $\rightarrow$} & \multirow{2}{*}{MModality $\uparrow$}\\
\cmidrule(lr){2-4}
& Top 1 & Top 2 & Top 3 &  \\
\midrule
Phys-GT & $0.481^{\pm 0.006}$ & $0.633^{\pm 0.008}$ & $0.722^{\pm 0.006}$ & $0.004^{\pm 0.000}$ & $3.401^{\pm 0.002}$ & $2.080^{\pm 0.011}$ & -  \\
\midrule
InterGen++ ~\citep{liang2024intergen} & $0.287^{\pm 0.007}$ & $0.439^{\pm 0.006}$ & $0.542^{\pm 0.006}$ & $0.943^{\pm 0.012}$ & $3.751^{\pm 0.007}$ & $\mathbf{2.044}^{\pm 0.006}$ & $\mathbf{2.482}^{\pm 0.003}$ \\
InterMask++ ~\citep{javed2024intermask}  & $0.156^{\pm 0.008}$ & $0.259^{\pm 0.009}$ & $0.339^{\pm 0.009}$ & $2.143^{\pm 0.035}$ & $4.027^{\pm 0.009}$ & $1.974^{\pm 0.011}$ & $1.939^{\pm 0.003}$ \\
PDP ~\citep{truong2024pdp} & $0.183^{\pm 0.006}$ & $0.291^{\pm 0.007}$ & $0.375^{\pm 0.009}$ & $1.268^{\pm 0.151}$ & $3.927^{\pm 0.010}$ & $1.954^{\pm 0.021}$ & $2.402^{\pm 0.013}$  \\
CLoSD ~\citep{tevet2024closd}   &  $0.244^{\pm 0.008}$ & $0.372^{\pm 0.009}$ & $0.470^{\pm 0.005}$ & $1.132^{\pm 0.020}$ & $3.827^{\pm 0.008}$ & $1.966^{\pm 0.008}$ & $1.474^{\pm 0.005}$ \\
\midrule
InterAgent (Ours) & $\mathbf{0.375}^{\pm 0.006}$ & $\mathbf{0.525}^{\pm 0.006}$ & $\mathbf{0.615}^{\pm 0.007}$ & $\mathbf{0.582}^{\pm 0.018}$ & $\mathbf{3.585}^{\pm 0.007}$ & $2.018^{\pm 0.008}$ & $1.903^{\pm 0.014}$ \\
\bottomrule
\end{tabular}}
\caption{Comparison of text-driven physics-based multi-agent control on the InterHuman ~\citep{liang2024intergen} test set, where $\pm$ indicates 95\% confidence interval and $\rightarrow / \uparrow / \downarrow$ means the closer / larger / smaller the better. \textbf{Bold} indicates best results.}
\label{table:compare_baseline_interhuman}
\end{table*}

\myparagraph{Dataset and evaluation metrics.}
We evaluate our model on the InterHuman~\citep{liang2024intergen} dataset.
It is a multi-human MoCap dataset with fine-grained text annotations, widely used as multi-human generation benchmark.
We follow the commonly adopted evaluation protocol in the text-to-motion literature, which consists of five metrics: (1) \textit{R-Precision} measures text–motion consistency via motion-to-text retrieval accuracy. (2) \textit{Fréchet Inception Distance (FID)} quantifies the distributional gap between generated and real motions. (3) \textit{Multimodal Distance (MMDist)} evaluates the alignment between text and its corresponding motion in the latent space. (4) \textit{Diversity} measures the variability across different generated motions. (5) \textit{Multimodality (MModality)} assesses the variation among motions generated from the same text prompt.
To compute these metrics, we train a motion encoder and a text encoder with a contrastive loss, mapping paired text–interaction samples to a shared latent space. 

\myparagraph{Implementation details.}
We implement InterAgent using 4 multi-stream DiT blocks, each with a latent dimension of 768. Every block is composed of 1 inter-stream fusion attention, 5 context-aware conditioning attention layers, and 3 independent linear projection layers. The prediction horizon is set to $m=4$, and the history buffer size to $h=364$. In practice, we uniformly downsample the first 360 frames in the history buffer to 12 frames and concatenate them with the most recent 4 frames, forming our history state input to the model. We adopt a frozen CLIP-ViT-L/14 model~\citep{radford2021learning} as the text encoder.  We further apply classifier-free guidance~\citep{ho2021classifier}, where 10\% of the CLIP embeddings are randomly masked to zero during training, and a guidance scale of 3.5 is used during sampling. All models are trained using AdamW~\citep{loshchilovdecoupled} optimizer with $\beta_1 = 0.9$, $\beta_2 = 0.999$, a weight decay of $2\times 10^{-5}$, and a maximum learning rate of $1\times 10^{-4}$. We employ a cosine learning rate schedule with $5K$ warm-up iterations and train the model for $80K$ iterations with a batch size of 256 on 8 NVIDIA GeForce RTX 4090 GPUs, taking approximately 12 hours in total.
\subsection{Comparison}
\begin{figure*}
    \centering

    \includegraphics[width=1.0\textwidth]{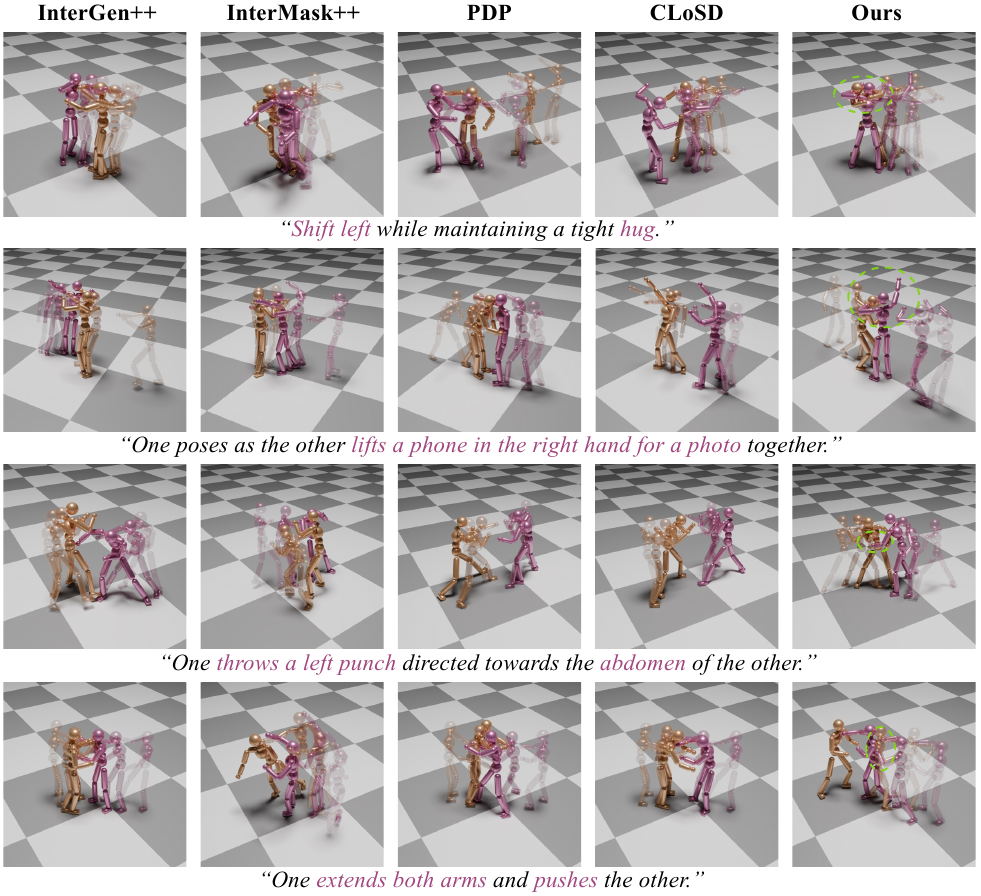}
    
    \caption{
    \textbf{Qualitative comparison.} Our InterAgent generates more coherent and natural multi-agent interactions that closely align with the given text commands, outperforming all baselines. 
    }
    \label{fig:qualitive_comparison}
\end{figure*}

We compare our method with four baselines. InterGen++ first generates kinematic motion trajectories from text using InterGen~\citep{liang2024intergen} and then employs tracking policy follow the generated trajectory to produce physically plausible motions. InterMask++ follows the same pipeline but replaces the kinematic generator with InterMask~\citep{javed2024intermask}. CLoSD~\citep{tevet2024closd} and PDP~\citep{truong2024pdp} are autoregressive text-to-motion models, which we extend to handle two-person interactions for comparison.
As shown in~\cref{table:compare_baseline_interhuman,fig:qualitive_comparison}, our model achieves superior text–motion alignment and overall realism compared to all baselines, while maintaining strong physical plausibility. Although slightly lower than InterGen~\citep{liang2024intergen} in Diversity and Multimodality, it consistently outperforms other methods across the remaining metrics. Qualitatively, kinematic–tracking approaches (InterGen++ and InterMask++) often fail to complete full motions or exhibit instability, and autoregressive extensions of CLoSD~\citep{tevet2024closd} and PDP~\citep{truong2024pdp} tend to miss fine-grained interaction details. In contrast, our method produces physically coherent and semantically faithful interactions—such as a tightly aligned \textit{“hug”} or a precisely targeted \textit{“punch towards the abdomen”}—demonstrating its ability to generate realistic and contextually accurate two-agent motions.
\begin{table}[t]
\centering
\resizebox{1.0\columnwidth}{!}{
\begin{tabular}{cccc}
\toprule
\multirow{2}{*}{Exteroception}  & {Num. of}  & {R-precision}& \multirow{2}{*}{FID $\downarrow$}\\

& DiT stream & (Top-3) $\uparrow$ & \\
\midrule
RS &  3   &  $0.588^{\pm 0.007}$ &   $0.676^{\pm 0.022}$  \\
\midrule
\multirow{3}{*}{FIG}           &  1   &   $0.523^{\pm 0.016}$   &  $0.828^{\pm 0.031}$     \\
            &  2   &      $0.608^{\pm 0.007}$   &  $0.662^{\pm 0.013}$    \\
           &  3   &       $0.612^{\pm 0.010}$   &  $0.634^{\pm 0.014}$     \\
\midrule
SIG (Ours)     &  3   &       $\mathbf{0.615}^{\pm 0.007}$   &  $\mathbf{0.582}^{\pm 0.018}$     \\
\bottomrule
\end{tabular}}
\caption{Quantitative evaluation of the stream number of multi-stream blocks and different exteroception representations.}
\label{table:ablation_main}
\end{table}

\subsection{Ablation Study}
\begin{figure*}
    \centering

    \includegraphics[width=\textwidth]{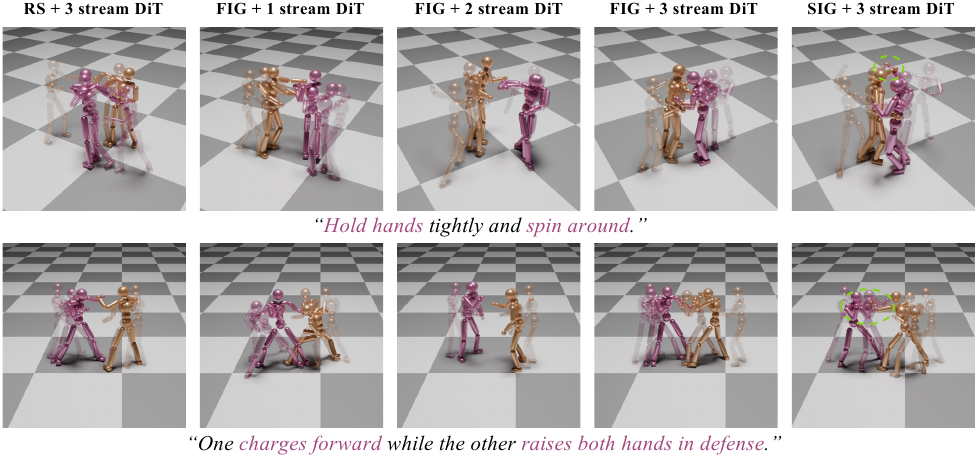}
    
    \caption{
    \textbf{Ablation study.} 
    We qualitatively evaluate how varying the number of streams in the multi-stream blocks and using different exteroception representations affect the quality and coherence of the generated multi-agent interactions.
    }
    \label{fig:ablation}
\end{figure*}

\myparagraph{Effect of different exteroceptions.}
To validate the effectiveness of our proposed SIG, we train variants with different exteroceptions (Rows 1, 4, and 5 in \cref{table:ablation_main}). Both FIG and SIG outperform RS in FID and R-precision, showing that interaction graphs yield more structured and informative inter-agent representations. Moreover, our SIG achieves lower FID and higher R-precision than FIG, indicating that its sparse attention effectively exploits the inherent sparsity of IG. Qualitative results (\cref{fig:ablation}, Columns 1, 4, and 5) further confirm that SIG produces more natural and accurate interactive behaviors.
\begin{table}[ht]
\centering
\resizebox{1.0\columnwidth}{!}{
\begin{tabular}{cccc}
\toprule
\multirow{2}{*}{IG Attention}  & \multirow{2}{*}{Sparsity Ratio}  & {R-precision}& \multirow{2}{*}{FID $\downarrow$}\\

&  & (Top-3) $\uparrow$ & \\
\midrule
non-sparse  &  -   &    $0.612^{\pm 0.010}$    & $0.634^{\pm 0.014}$    \\
\midrule
\multirow{4}{*}{edge-based}  &  3/4  &   $0.609^{\pm 0.006}$    & $0.617^{\pm 0.026}$    \\
  &  1/2  &   $\mathbf{0.615}^{\pm 0.007}$    & $\mathbf{0.582}^{\pm 0.018}$    \\
  &  1/4   &  $0.602^{\pm 0.010}$    & $0.624^{\pm 0.026}$   \\
  &  1/8  &  $0.601^{\pm 0.008}$    & $0.643^{\pm 0.020}$     \\
\midrule
\multirow{4}{*}{joint-based} &  3/4  &   $0.608^{\pm 0.010}$    & $0.615^{\pm 0.022}$     \\
 &  1/2  &   $\mathbf{0.615}^{\pm 0.011}$    & $0.619^{\pm 0.021}$     \\
 &  1/4  &   $0.607^{\pm 0.010}$    & $0.591^{\pm 0.020}$    \\
 &  1/8   &  $0.609^{\pm 0.012}$    & $0.603^{\pm 0.011}$     \\
\bottomrule
\end{tabular}}
\caption{Quantitative analysis of the IG attention and the effect of different sparsity ratios.}
\label{table:ablation_sparse_attn}
\end{table}

\myparagraph{Effect of multi-stream DiT block.}
To evaluate the multi-stream design of Inter-DiT, we compare it with a single-stream DiT variant that predicts proprioception, exteroception, and action jointly, and a dual-stream variant that predicts state (proprioception and exteroception) and action separately. As shown in \cref{table:ablation_main} (Rows 2, 3, and 4) and \cref{fig:ablation} (Columns 2, 3, and 4), our triple-stream Inter-DiT consistently outperforms both alternatives across evaluation metrics and qualitative visualizations, demonstrating the effectiveness of the decoupled architecture. 

\myparagraph{Effect of IG attention and sparsity ratios.}
We further investigate different sparsity strategies by varying IG attention and sparsity ratios, with edge-based targeting individual joint-to-joint edges and joint-based treating all edges incident to a joint as a single unit. Results (\cref{table:ablation_sparse_attn}) show that edge-based IG attention with a 1/2 ratio achieves the best performance, indicating that selectively removing less relenvant edges more effectively captures inter-agent dependencies while mitigating redundancy, thus improving representation efficiency without sacrificing key relational cues.

\begin{figure}[htpb]
    \centering

    \includegraphics[width=\columnwidth]{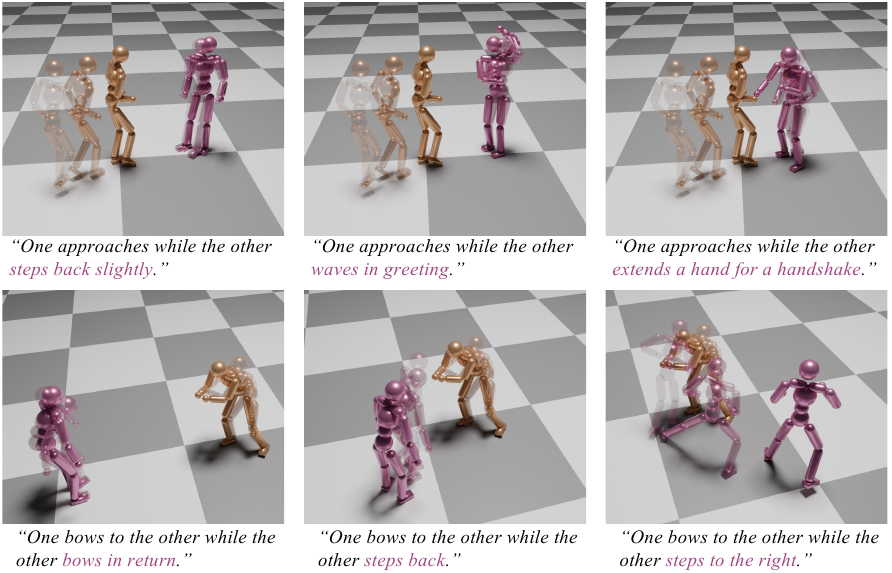}
    
    \caption{
    \textbf{Reactive humanoid control.} 
    We demonstrate reactive multi-agent control, in which one agent’s motion (gold) is fixed while the other (pink) generates text-conditioned reactions.
    }
    \label{fig:reaction}
\end{figure}

\subsection{Reactive Humanoid Control}
InterAgent enables reactive humanoid control without the need for retraining by incorporating an inpainting mechanism~\citep{xu2025moreact,liang2024intergen,tevet2022human} during inference. Specifically, we fix the behaviors of one agent via replay, and at each timestep $t$ of the denoising process, after predicting the clean trajectory, we overwrite the fixed agent’s predicted proprioception with its ground truth to guide the generation of responsive behaviors. The visualization results are presented in \cref{fig:reaction}, where the golden humanoid denotes the fixed behaviors and the pink one represents the text-conditioned generated response. The results demonstrate that our model is capable of synthesizing a diverse repertoire of reaction that are closely adhere to the given textual commands.

\section{Conclusion} \label{sec:conclu}

In this paper, we presented InterAgent, a novel physics-based framework for text-driven multi-agent humanoid control. By introducing multi-stream DiT architecture, interaction graph exteroception, and the tailored edge-based sparse attention mechanism, our approach effectively models fine-grained inter-agent dependencies while preserving physically plausible and dynamically coherent behaviors. Comprehensive quantitative and qualitative evaluations demonstrate that InterAgent consistently outperforms all baselines in coordinated and physically realistic text-driven multi-agent humanoid control. We firmly believe it can establish a solid foundation for future research in physics-based multi-agent humanoid control and open avenues for broad applications in gaming, VR/AR, and autonomous robots.

\section{Acknowledgement}
This work was supported by Shanghai Local College Capacity Building Program (23010503100),  NSFC (No.62406195, W2431046, W2431046,),Shanghai Frontiers Science Center of Human-centered Artificial Intelligence (ShangHAI), MoE Key Laboratory of Intelligent Perception and Human-Machine Collaboration (ShanghaiTech University), the Shanghai Frontiers Science Center of Human-centered Artificial Intelligence, HPC Platform and Core Facility Platform of Computer Science and Communication of ShanghaiTech University and Shanghai Engineering Research Center of Intelligent Vision and Imaging.

{
    \small
    \bibliographystyle{ieeenat_fullname}
    \bibliography{main}
}
\clearpage
\setcounter{page}{1}
\maketitlesupplementary

\section{Demo Video}
Beyond the qualitative snapshots, we provide a demo video offering more detailed visualizations, further showcasing the effectiveness of our approach. 

\section{Additional Implementation Details}

\myparagraph{Interaction tracking policy.}
To ensure the collected data accurately captures the ongoing interactions among agents, we integrate the interaction graph reward~\citep{zhang2023simulation} into the training of the tracking policy. Specifically, during the data collection phase, we define each edge of the interaction graph as $\bm{e}_{ij}= (\bm{p}_{ij}, \bm{v}_{ij})$, where $\bm{p}_{ij}$ and $\bm{v}_{ij}$ denote the relative position and relative velocity between joint $i$ and joint $j$, respectively. Based on this edge representation, we further define the discrepancy for each pair of edges as: 
\begin{equation}
    d_{pos} = \sum_{i,j \in E} \omega_{ij} ||\bm{p}^{ref}_{ij}-\bm{p}^{sim}_{ij}||,
\end{equation}
\begin{equation}
    d_{vel} = \sum_{i,j \in E} \gamma_{ij} ||\bm{v}^{ref}_{ij}-\bm{v}^{sim}_{ij}||,
\end{equation}
where the superscripts $ref$ and $sim$ correspond to the reference motion and simulation motion, $\omega_{ij}$ and $\gamma_{ij}$ are edge weighting functions detailed in~\citet{zhang2023simulation}. Accordingly, we formulate the interaction graph reward as follows:
\begin{equation}
    r_{ig} = \exp{(-\lambda_{pos} * d_{pos} -\lambda_{vel} * d_{vel})},
\end{equation}
The root reward is defined as:
\begin{equation}
    r_{root} = \exp{(-\lambda_{root} *  ||\bm{x}^{ref}_{root}-\bm{x}^{sim}_{root}||)},
\end{equation}
where the $\lambda_{pos}$, $\lambda_{vel}$ and $\lambda_{root}$ denote the term sensitivities, $\bm{x}_{root}$ represents the position, velocity, orientation and angular velocity of the root joint. Therefore, our reward function is then calculated as:
\begin{equation}
    r = r_{ig} * r_{root} .
\end{equation}

\myparagraph{Reactive humanoid control.}
To enable reactive behaviors, we embed an inpainting mechanism within the inference process. For a given command $\bm{c}$, we fix the behaviors of a humanoid via replaying its ground truth. At each denoising step $t$, upon the model predicting the clean behaviors $\hat{\bm{X}} = [\hat{\bm{x}}_{p_1}, \hat{\bm{x}}_{e_1}, \hat{\bm{x}}_{a_1}, \hat{\bm{x}}_{p_2}, \hat{\bm{x}}_{e_2}, \hat{\bm{x}}_{a_2}]$, we substitute the generated proprioception of the fixed humanoid with its ground truth equivalents. For concreteness, we assume this substitution targets $\hat{\bm{x}}_{p_1}$ here, yielding $\hat{\bm{X}'} = [{\bm{x}}_{p_1}, \hat{\bm{x}}_{e_1}, \hat{\bm{x}}_{a_1}, \hat{\bm{x}}_{p_2}, \hat{\bm{x}}_{e_2}, \hat{\bm{x}}_{a_2}]$, which is then used to advance to the subsequent sampling process. This ensures that the generated interaction dynamics strictly constrained by the fixed humanoid's behaviors. Empirically, we refrain from replacing generated actions with ground truth counterparts, as the intrinsic randomness of simulation environment makes it difficult for the humanoid to maintain balance when the actions in collected dataset are directly deployed.

\section{Additional Experimental Results}
In this section, we introduce experimental results that are not included in the main paper due to space limit.

\begin{figure}[htpb]
    \centering

    \includegraphics[width=\columnwidth]{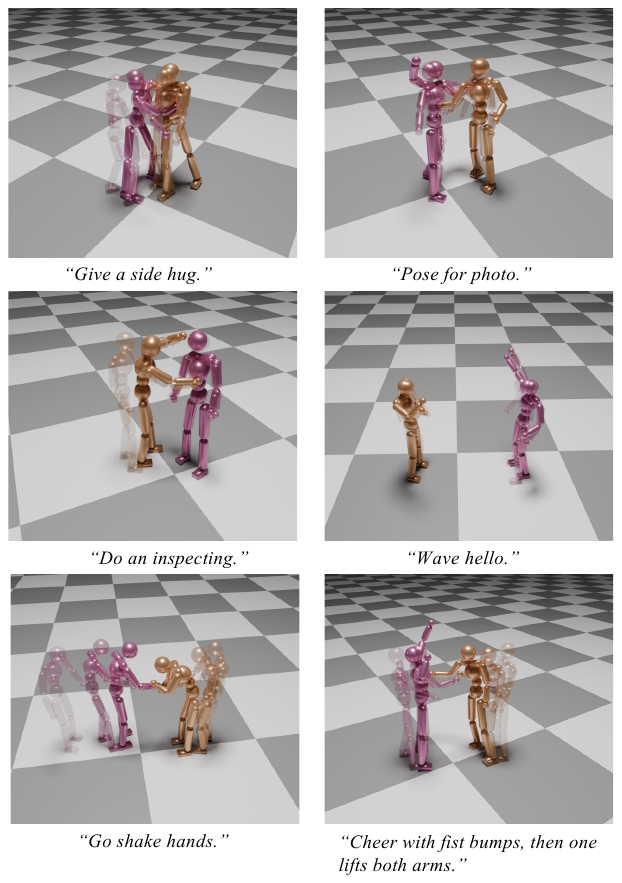}
    
    \caption{
    \textbf{Qualitative results.} 
    }
    \label{fig:addition_quality}
\end{figure}

\myparagraph{Qualitative results.} As shown in~\cref{fig:addition_quality}, We show additional qualitative results generated by our model given various textual commands.

\myparagraph{Quantitative physical analysis.}\label{sec:9}
To evaluate the physical correctness, we analyze three widely used metrics~\citep{yuan2023physdiff,wu2025uniphys}: \textit{Floating} (measuring vertical drift), \textit{Skating} (indicating horizontal sliding), and \textit{Jerk} (quantifying the smoothness of the motion). Penetration is not reported herein, as it is negligible across all physics-based methods.  As shown in~\cref{table:physics_metric}, InterMask++~\citep{javed2024intermask} exhibits significantly poor performance in floating and jerk metrics. Such deficiencies make it inherently challenging for humanoids to maintain balance in physical environments, let alone execute complex interactive behaviors. Consequently, such physical inconsistencies directly contribute to its poor FID and R-precision scores. In contrast, Our InterAgent demonstrates competitive performance across all three physical metrics, closely approaching the ground truth (Phys-GT) and outperforming most baseline methods. It strikes a strong balance between miniming floating, controlling skating, and reducing jerk, making it robust for text-driven physics-based multi-agent humanoid control. 
\begin{table}[t]
\centering
\resizebox{1.0\columnwidth}{!}{
\begin{tabular}{cccc}
\toprule
  & {Floating $\downarrow$}  & {Skating $\downarrow$}& {Jerk $\downarrow$}\\

& $[mm]$ & $[mm]$ & $[mm/{frame}^3]$ \\
\midrule
Phys-GT                                 &  49.92   &  $4.07\times10^{-6}$ &   16.24  \\
\midrule
InterGen++~\citep{liang2024intergen}   &  53.35   &   $1.24 \times 10^{-3}$   & 12.56     \\
InterMask++~\citep{javed2024intermask} &  151.34  &   $4.29 \times 10^{-4}$  &  39.15    \\
PDP~\citep{truong2024pdp}              &  $\underline{49.84}$   &   $7.49 \times 10^{-4}$   &  
$\mathbf{1.98}$ \\
CLoSD~\citep{tevet2024closd}           &  $\mathbf{48.64}$   &   $\mathbf{2.62 \times 10^{-5}}$   &  
29.07 \\
\midrule
InterAgent (Ours)                      &  49.85   &   $\underline{2.81 \times 10^{-4}}$ &  $\underline{2.69}$  \\
\bottomrule
\end{tabular}}
\caption{Quantitative evaluation of physical correctness. \textbf{Bold} and \underline{underline} indicate the best and the second
 best result.}
\label{table:physics_metric}
\end{table}

\section{Discussion}
\begin{figure}[htpb]
    \centering

    \includegraphics[width=\columnwidth]{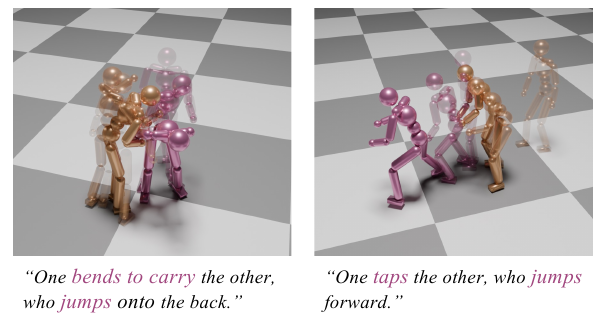}
    
    \caption{
    \textbf{Failure cases.} Our method struggles with highly dynamic behaviors like jumping.
    }
    \label{fig:failure}
\end{figure}

\myparagraph{Limitations and failure cases.}
As shown in~\cref{fig:failure}, our method performs well for most cases but still struggles with more dynamic behaviors like jumping. This issue mainly duo to model's bias toward smooth transitions, which conflicts with the significant instantaneous dynamics of jumping (explosive push-off, mid-air balance, landing impact). A potential solution could involve a dynamic physical constraint module tailored to high-dynamic behaviors. Additionally, augmenting training data with annotated high-dynamic sequences may further enhance the model’s adaptability while preserving its steady-state performance.

\myparagraph{Discussion and future work.}
While InterAgent demonstrates strong capability in generating coordinated, physically plausible, and text-consistent multi-agent behaviors, several aspects of the framework present valuable opportunities for further advancement. Firstly, the current text interface describes high-level motion intent but does not explicitly reason about long-horizon task structure, role assignment, or interactive strategies. Integrating a higher-level planning module, or combining large language models with our physics-based controller, could enable richer collaborative behaviors and more natural multi-turn interactions. Secondly, our model is trained with a fixed number of agents, and the computational cost grows with the number of entities due to pairwise relational modeling. Extending the framework toward scalable multi-agent coordination—possibly through hierarchical grouping, clustering of interaction patterns, or dynamic neighborhood selection—will be an essential step toward deployment in complex multi-agent interaction environments.
Finally, deploying InterAgent beyond simulation remains an exciting challenge, applying our framework to real humanoid robots or VR/AR avatars will require addressing sim-to-real transfer, real-time inference, and robust perception under noisy sensory inputs. We envision InterAgent serving as a building block for future interactive embodied systems, enabling fluid multi-agent coordination in entertainment, robotics, and immersive virtual environments.

\end{document}